\documentclass[letterpaper, 10 pt, conference]{ieeeconf}  

\IEEEoverridecommandlockouts                              

\usepackage{epsfig}
\usepackage{graphicx}
\usepackage{amsmath}
\usepackage{amssymb}
\usepackage{color}
\usepackage{booktabs}
\usepackage{colortbl}
\usepackage{url}

\title{\LARGE \bf
Making Robots Draw A Vivid Portrait In Two Minutes
}

\author{Fei Gao$^{1,2}$, Jingjie Zhu$^{1,2}$, Zeyuan Yu$^{1}$, Peng Li$^{1,\dag}$, Tao Wang$^{1,3}$, \textit{Senior Member, IEEE} 
\thanks{*This work was supported in part by the National Natural Science Foundation of China under Grants 61971172 and 61971339, and in part by the China Post-Doctoral Science Foundation under Grant 2019M653563.}
\thanks{$^{1}$Intelligent Computing Research Center, Advanced Institute of Information Technology (AIIT), Peking University, Hangzhou 311215, China.}%
\thanks{$^{2}$School of Computer Science and Technology, Hangzhou Dianzi University, Hangzhou 310018, China.}
\thanks{$^{3}$Department of Computer Science and Technology, Peking University, Beijing 100871, China.}%
\thanks{$^\dag$Corresponding Author. Peng Li: {\tt\small pli@aiit.org.cn}}
}

\begin{document}

\maketitle
\thispagestyle{empty}
\pagestyle{empty}

\begin{abstract}

Significant progress has been made with artistic robots. 
However, existing robots fail to produce high-quality portraits in a short time. 
In this work, we present a drawing robot, which can automatically transfer a facial picture to a vivid portrait, and then draw it on paper within two minutes averagely. 
At the heart of our system is a novel portrait synthesis algorithm based on deep learning. 
Innovatively, we employ a self-consistency loss, which makes the algorithm capable of generating continuous and smooth brush-strokes.  
Besides, we propose a componential sparsity constraint to reduce the number of brush-strokes over insignificant areas.
We also implement a local sketch synthesis algorithm, and several pre- and post-processing techniques to deal with the background and details. 
The portrait produced by our algorithm successfully captures individual characteristics by using a sparse set of continuous brush-strokes.
Finally, the portrait is converted to a sequence of trajectories and reproduced by a 3-degree-of-freedom robotic arm. 
The whole portrait drawing robotic system is named AiSketcher.
Extensive experiments show that AiSketcher can produce considerably high-quality sketches for a wide range of pictures, including faces in-the-wild and universal images of arbitrary content.  
To our best knowledge, AiSketcher is the first portrait drawing robot that uses neural style transfer techniques. 
AiSketcher has attended a quite number of exhibitions and shown remarkable performance under diverse circumstances.

\end{abstract}

\section{Introduction}
\label{sec:intro}


The ability of robots to create high-quality artworks is popularly considered a benchmark of progress in Artificial Intelligence (AI). 
Reaserchers have contributed great efforts to develop artistic robots, which can draw sketch portraits \cite{Gao2019}, colourful pictures \cite{IROS2016Luo}, watercolors \cite{Watercolour}, etc. Some of them have attended various exhibitions and shown impressive performance \cite{RobotArt, Robotlab}. 
Drawing colourful images typically costs a robot several hours to finish it \cite{Watercolour}. In contrast, a portrait typically contains a parse set of graphical elements (e.g., lines) and can be finished in a short time. Portrait drawing robots thus allow active interactions between robots and normal consumers. 

In recent years, portrait drawing robots has attracted a lot of attention. 
Existing robots typically produce portraits by means of low-level image processing \cite{Gao2019}. They are unable to draw high-quality portraits, especially for faces in-the-wild, i.e. faces presenting extreme poses, expressions, or occlusions, etc. In addition, portraits have a highly abstract style. Even for artists, portraits drawing relies on professional training and experience \cite{APDrawingGAN}. It is still a complicated and elaborate task for robots to draw high-quality portraits.

The other critical challenge is to balance the \textit{vividness} of a sketch and the \textit{time-budget} for drawing it.
Vivid sketches can please human users, while limited time-budget would avoid their impatience. 
However, the vividness and the drawing time contradict each other in practice. 
Vividness is correlated with details about characteristics in a given face. 
Generating more details in a portrait generally improves the vividness, but prolongs the drawing process. 
Conversely, a portrait with a sparse set of elements can be quickly drawn but may fail to capture  individualities. 


To address the above challenges, in this work, we develop a novel portrait drawing robot by means of deep neural networks \cite{CNN2012}. The proposed method is inspired by the great success of neural style transfer (NST) in creating various types of artworks \cite{Jing2019Neural}. To our best knowledge, no drawing robots have been developed based on these NST algorithms. Besides, preliminary experiments show that these algorithms cannot produce continuous and smooth brush-strokes. 

In this work, we propose a novel portrait synthesis algorithm for drawing robots. Specially, we propose two novel objectives for generating sparse and continuous brush-strokes. 
First, we enforce the algorithm capable of reconstructing real sketches (of arbitrary content) by using a self-consistency loss. This constraint considerably improves the continuity and realism of synthesised brush-strokes. 
Second, we apply a sparsity constraint to regions which are insignificant for characterizing individualities. 
The corresponding componential sparsity loss significantly reduces the number of brush-strokes without degrading the quality of a portrait. 
Here, we use face parsing masks to represent facial components. 
We also implement a specific local synthesis algorithm as well as several pre- and post-processing techniques to deal with the background and details. 


Finally, we develop a drawing robot, namely AiSketcher, by using a 3-degree-of-freedom (DOF) robotic arm. 
The synthesised sketch is converted into a sequence of trajectories that the robot reproduces on paper or other flat materials. 
In the robotic system, we implement several human-robot interaction techniques to facilitate uses of AiSketcher. 
Extensive experimental results show that AiSketcher outperforms previous state-of-the-art. Besides, AiSketcher works well over a wide range of images, including faces in-the-wild and universal images of arbitrary content.  
AiSketcher has attended a quite number of exhibitions and shown remarkable as well as robust performance under diverse circumstances.

In summary, we make the following contributions:
\begin{itemize}
\item AiSketcher is the first portrait drawing robot that uses neural style transfer techniques. 
\item We propose a self-consistency loss to make the algorithm produce continuous and realistic brush-strokes.
\item We propose a componential-sparsity loss to balance the vividness and time-budget. 
\item Our portrait synthesis algorithm doesn't need a large-scale photo-sketch pairs for training. 
\item Our robotic system works well for faces in-the-wild and a wide range of universal images.
\end{itemize}

The paper is organized as follows: Section \ref{sec:relatedwork} provides a brief description of related work. Section \ref{sec:arch} presents an overview on the whole system. In Section \ref{sec:fss} and Section \ref{sec:robotic}, the portrait synthesis algorithms and the robotic system are detailed, respectively. Extensive experiments are presented in Section \ref{sec:exp}. Finally, we conclude this work in Section \ref{sec:conclusion}.

\section{Related Work}
\label{sec:relatedwork}

In this section, we briefly introduce some related work, including portrait drawing robots, face sketch synthesis, deep neural style transfer, and generative adversarial network.  

\subsection{Portrait Drawing Robots}
\label{ssec:rel-robots}

A number of efforts have been devoted to develop portrait drawing robots. 
For example, Calinon et al. \cite{Calinon2005} develop a humanoid robot drawing binary facial pictures. 
Lin et al. \cite{Lin2009} use center-off algorithm and median filtering to generate a portrait, and then refine it based on low-level textural features. The resulting portraits are unduly simple and roughly present geometric outlines of human faces. 
Mohammed et al. \cite{Mohammed2013} use edge detection and counter tracing to synthesis sketches and plan the path. 
Additionally, Lau et al. \cite{Lau2012} explore a portrait robot by means of point-to-point drawing instead of line-drawing. 
Tresset et al. \cite{Paul2013} extract Gabor kernels to extract salient lines from a detected face, and finish drawing by means of visual feedback. 
Recenlty, Ye et al. \cite{Ye2017} use deep learning based method for face detection and extraction, and then use the holistically-nested edge detection (HED) algorithm \cite{Shen2015HED} to extract head edges as the portrait.



Since key-points represent facial geometry, Xue and Liu \cite{Xue2017} use these key-points and a mathematical morphology algorithm to generate a primal sketch. Then, they generate some steaks to represent hairs based on textural features. Similarly, Gao et al. \cite{Gao2019} use key-points to synthesis an outline portrait, and then use random points and lines to represent eyebrows and hair. 
The portraits produced by these methods typically present stiff steaks. Besides, these methods rely heavily on key-points detection algorithms, which currently don't work well for faces in-the-wild. 
Finally, the generated strokes are not in an artistic style. 

\subsection{Deep Neural Style Transfer}
\label{ssec:rel-ist}
Neural Style Transfer (NST) means using Convolutional Neural Networks (CNNs) to render a content image in a target style \cite{Jing2019Neural}.  
The problem of style transfer is closely related to texture synthesis and transfer \cite{Efros1999}. Early approaches typically rely on low-level statistics. Recently, Gatys et al. \cite{Gatys} first use CNNs to solve this problem and have lead to a trending developments both in academic literature and industrial applications \cite{Jing2019Neural}. Inspired by the success of NST, we develop a novel portrait synthesis algorithm in this work. 

\subsection{Generative Adversarial Networks}
\label{ssec:rel-gan}
Recently, Generative Adversarial Networks (GANs) \cite{GAN,Pix2Pix} have show inspiring results in generating facial pictures, natural images, facial sketches, paintings etc. 
In GAN, a neural network works as generator $G$, which learns to transfer a source image $x$ to a target image $y$. The other network works as discriminator $D$, which aims at distinguishing $y$ with the generated image $G(x)$ \cite{Pix2Pix}. $G$ and $D$ are alternately optimized in an adversarial manner. Finally, $G$ would be capable of producing images in the style of $y$.
There have been a huge number of algorithms and applications of GANs. Please refer to  \cite{Gui2020ReviewGAN} for a comprehensive study. 

Recently, Yi et al. propose an APDrawingGAN model for generating artistic portrait drawing \cite{APDrawingGAN}. However, the resulting sketches present too many dilated details and black areas, which would significantly increase the drawing time of a robot. Besides, APDrawingGAN cannot produce high-quality portraits for faces in-the-wild. 
Most recently, Wang et al. \cite{Wang2017RoboCoDraw} propose an AvatarGAN, based on CycleGAN \cite{CycleGAN}, for generating avatars for human faces and develop a drawing robot. Differently, we develop a robotic system for drawing \textit{realistic-style} portraits based on NST and GAN. 

\subsection{Facial Sketch Synthesis}
\label{ssec:rel-fss}
Another close topic to our work is face sketch synthesis (FSS). 
FSS generally means synthesizing a pencil-drawing sketch based on an input face photo \cite{WangIJCV}. Recently, researchers develop some NST \cite{ChenNST} or GAN-based \cite{CAGAN} methods, which can produce fantastic pencil-drawing sketches. 
However, pencil-drawing sketches typically consume a long time to draw. They are thus unsuitable for real-time drawing robots. 

   \begin{figure*}[thpb]
      \centering
      \includegraphics[width=1\linewidth]{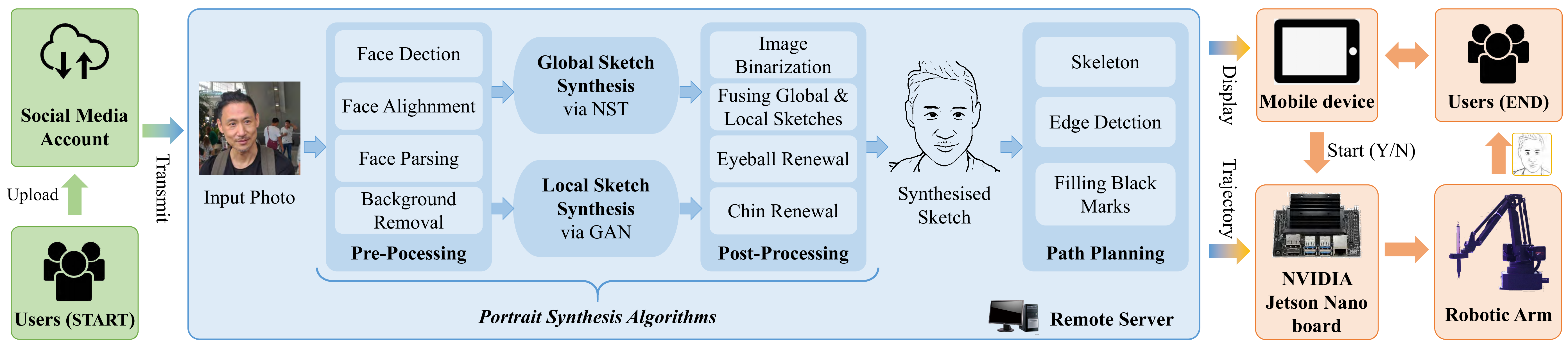}
      \caption{Overview of our portrait drawing robotic system, AiSketcher.}
      \label{fig:pipeline}
   \end{figure*}

\section{Architecture of The System}
\label{sec:arch}

In this section, we overview the architecture of the robotic portrait drawing system. 
It is composed of both software (e.g. the algorithms for portrait synthesis, Section \ref{sec:fss}) and hardware components (e.g. the robotic arm and remote server, Section \ref{sec:robotic}). Fig. \ref{fig:pipeline} shows an overview on the system.

First, a facial picture is uploaded by users through a social media account and fed into the portrait synthesis algorithms, as explained in Section \ref{sec:fss}. 
Afterwards, the resulting portrait is converted to a sequence of trajectories, by the path planning module. 
Finally, the list of commands is deployed on a NVIDIA Jetson Nano board to control the robotic arm, which reproduces the portrait on paper or other flat materials. 
Additionally, we provide a mobile device for users to choose a preferred portrait from a number of alternatives, and to start the drawing procedure.



\section{Algorithms for Portrait Synthesis}
\label{sec:fss}

The pipeline of our portrait synthesis algorithms is as shown in Fig. \ref{fig:pipeline}.
Given a facial picture $x$, we first pre-process it by means of face detection, face alignment, face parsing, and background removal. 
The resulting picture is denoted by $\tilde{x}$. 
Afterwards, we feed $\tilde{x}$ into a global sketch synthesis algorithm, which produces a primary portrait. 
Additionally, we use a local sketch synthesis algorithm to specifically synthesis eyebrows. 
We then fuse the globally and locally synthesised sketches, and deal with details, in the post-processing stage.

   \begin{figure}[thpb]
      \centering
      \includegraphics[width=1\linewidth]{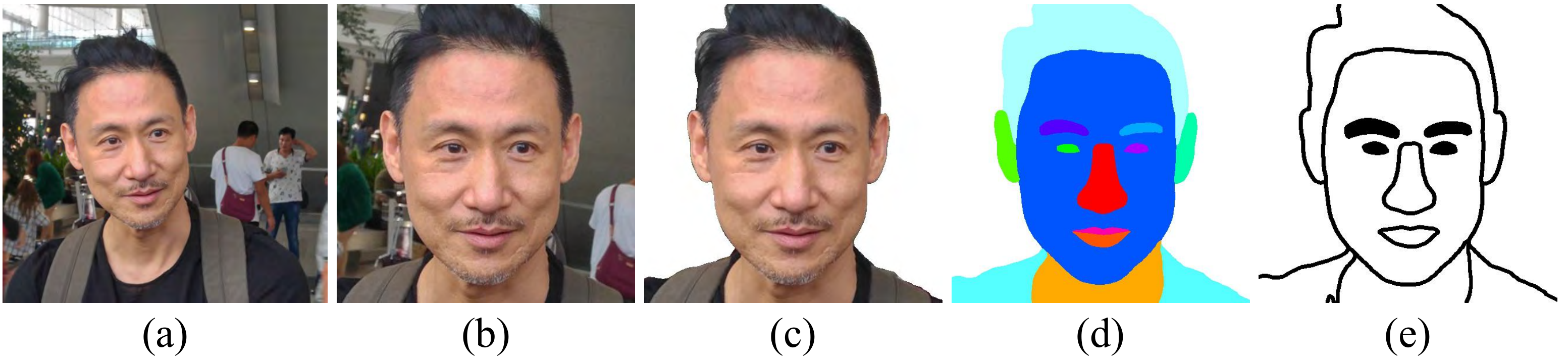}
      \caption{Illustration of image pre-processing. (a) Input picture, (b) aligned picture, (c) background removal, (d) face parsing mask $M$, and (e) compositional sparsity mask $M'$.}
      \label{fig:preprocmask}
   \end{figure}

\subsection{Pre-processing}
\label{ssec:prepoc}
We use the following approaches to align an input picture and to deal with the background. 

\textbf{Face Detection.} First, we use the \texttt{dlib} toolbox in OpenCV \cite{opencv} to detect the face in a given image, as well as 68 key-points in the face. In default settings, if more than one faces are detected, we choose the largest one. If no face is detected, we directly feed the original image into the global sketch synthesis algorithm to produce the \textit{final} output. 

\textbf{Face Alignment.} Afterwards, we geometrically align the facial picture by affine transformation, relying on centers of eyes. The aligned image is automatically cropped to $512 \times 512$ pixels (1:1) or $512 \times 680$ pixels (3:4), according to the size of the original picture. Fig. \ref{fig:preprocmask}b shows an aligned example. 

\textbf{Face Parsing.} Then, we use face parsing to represent facial components. Specially, we use the BiSeNet \cite{yu2018bisenet} trained on the CelebAMask-HQ dataset \cite{MaskGAN} \footnote{Online available: https://github.com/zllrunning/face-parsing.PyTorch} to segment an aligned picture into 19 components, including background, hair, two eyebrows, two eyes, nose, two lips, neck, clothes, etc. Fig. \ref{fig:preprocmask}d illustrates a parsing mask $M$, where facial components are distinguished by colors.

\textbf{Background Removal.} Finally, based on the parsing mask, we replace the background with white pixels (Fig. \ref{fig:preprocmask}c), so no elements would be generated in the background. 

\subsection{Global Sketch Synthesis}
\label{ssec:global}

\begin{figure}[thpb]
      \centering
      \includegraphics[width=1\linewidth]{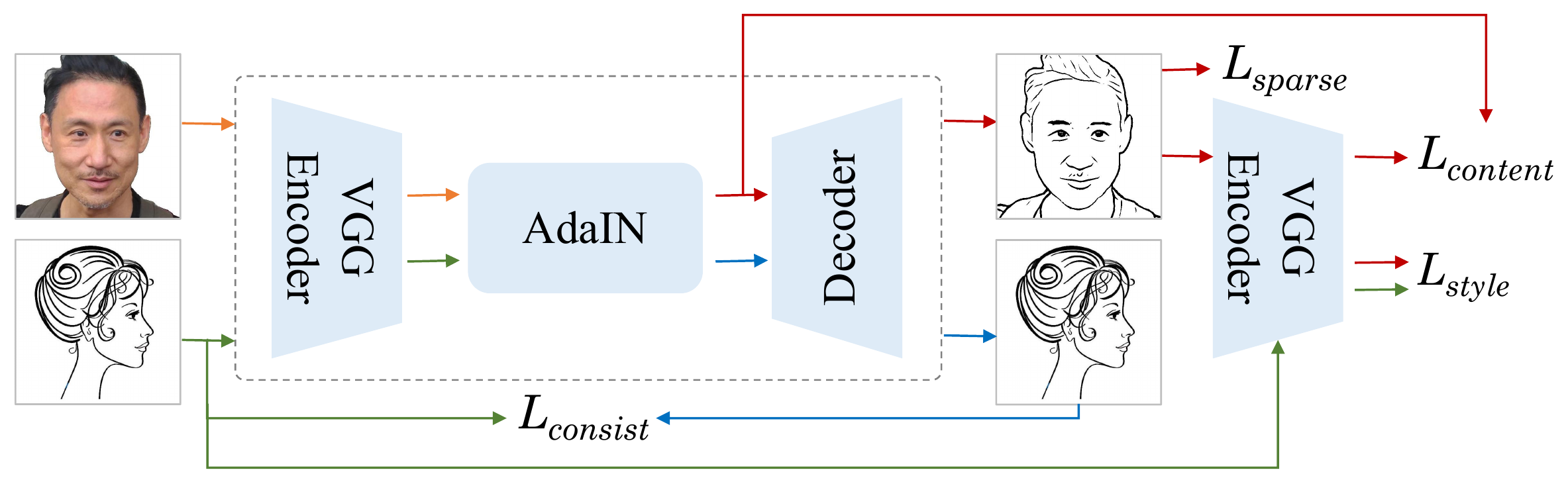}
      \caption{Pipeline of the global sketch synthesis algorithm. }
      \label{fig:adain}
   \end{figure}

We first develop a global sketch synthesis algorithm to transfer the facial picture $\tilde{x}$ to a sketch. 
Here, we adopt the algorithm propoed in \cite{AdaIN} as our base model, due to its inspiring performance for arbitrary style transfer. 
However, this algorithm cannot generalize sparse and continuous brush-strokes for sketch synthesis. 
We thus propose two novel loss functions, i.e. the \textit{Self-consistency Loss} and \textit{Compositional Sparsity Loss}, to boost the quality of sketch synthesis. 
The framework of our global sketch synthesis network is as shown in Fig. \ref{fig:adain}. 
Details will be presented bellow. 

\textbf{Network Architecture.} 
The global synthesis algorithm takes the pre-processed face photo $\tilde{x}$ and a style image $s$ as inputs, and outputs a synthesised sketch. Here, we use a real sketch $s$ (of arbitrary content) as the style image. 

We use the same network architecture as that used in \cite{AdaIN}. Specially, we use the first few layers of a pre-trained VGG-19 \cite{VGG} as an encoder $f$.  
Both the content and style images are fed into $f$ and converted to feature maps: $f(\tilde{x})$ and $f(s)$. 
Afterwards, an AdaIN layer produces the target feature maps $t$ by:
$$
t = \sigma(f(s)) \frac{f(\tilde{x})-\mu(f(\tilde{x}))}{\sigma(f(\tilde{x}))}+\mu(f(s)), \eqno{(1)}
$$
where $\mu(\cdot)$ and $\sigma(\cdot)$ denote the mean and standard variance of features. 
Finally, $t$ is fed into a decoder $g$, generating a sketch portrait $T(\tilde{x}, s)$:
$$
T(\tilde{x}, s) = g(t)  \eqno{(2)}
$$
$g$ includes nine convolutional layers and three nearest up-sampling layers. Reflection padding is applied to each layer. 

In the implementation, $f$ is fixed and $g$ is randomly initialized. We use a pre-trained VGG-19 to compute the following four loss functions for training $g$.

\textbf{Content Loss.} 
The content loss encourages a synthesised sketch $T(\tilde{x},s)$ has the same content as the input facial picture $\tilde{x}$. It is formulated as the Euclidean distance between the target features $t$ and the features of the portrait $T(\tilde{x}, s)$: 
$$
L_{content} =  \Vert f(T(\tilde{x}, s)) - t \Vert _2  \eqno{(3)}
$$

\textbf{Style Loss.} 
In AdaIN, the style is represented by the mean and standard deviation of deep features. The style loss is correspondingly formulated as:
$$
L_{style} = \sum^L_{i=1} \Vert \mu(\phi_i(T(\tilde{x}, s))) - \mu(\phi_i(s)) \Vert _2
$$
$$
 + \sum^L_{i=1} \Vert \sigma(\phi_i(T(\tilde{x}, s))) - \sigma(\phi_i(s)) \Vert _2  \eqno{(4)}
$$
where $\phi_i(\cdot)$ denotes the outputs over the $i$-th layer in VGG-19. In the experiments, we use $relu1\_1$, $relu2\_1$, $relu3\_1$, and $relu4\_1$ layers to compute the style loss. .

\textbf{Self-consistency Loss.} 
We argue that the learned mapping function $T(\cdot)$ should be self-consistent: For a real sketch image $s$, the algorithm should be able to reconstruct it, i.e. $T(s,s) = g(f(s)) \approx s$. The corresponding self-consistency loss is formulated as:
$$
L_{consist} = \Vert T(s,s) - s \Vert_2  \eqno{(5)}
$$   
Experiments show that the self-consistency loss significantly improves the continuity of synthesised brush-strokes.

\textbf{Compositional Sparsity Loss.} 
Finally, we employ a compositional sparsity loss to constrain the network producing a sketch with a sparse set of brush-strokes. High-quality portraits mainly present brush-strokes for primary components or geometric boundaries. 
We therefore merely add a sparse constraint to relatively unnecessary regions, such as the background, hair, skin, and body. 
The corresponding compositional sparsity loss is formulated as: 
$$
L_{sparse} = \Vert M^{'} \odot ( 1 - T(\tilde{x}, s)) \Vert_1  \eqno{(6)}
$$   
where $\odot$ denotes the element-product operation. $M'$ is a binary sparsity mask derived from the parsing mask $M$. Specially, $M'$ is of the same size as $M$ and $\tilde{x}$. In $M'$, pixels corresponding to boundaries, eyes, eyebrows, and lips are assigned 0. All the rest positions are assigned 1 in $M'$, and encouraged to present white color in $T(\tilde{x},s)$. Since the predicted parsing mask might be imprecise, we slightly enlarge the black areas in $M'$ by image erosion.
The final compositional sparsity mask is as illustrated in Fig. \ref{fig:preprocmask}e. 

Using a global sparsity, i.e. $M'$ is full of 1, also leads to a parse set of elements in $T(\tilde{x},s)$. However, it reduces the elements about individual characteristics as well. Corresponding experiments will be presented in Section \ref{ssec:ablation}.

\textbf{Full Objective.} 
Our full objective is:
$$
L = \lambda_1 L_{content} + \lambda_2 L_{style} + \lambda_3 L_{consist} + \lambda_4 L_{sparse}   \eqno{(7)}
$$
where, $\lambda_1,\lambda_2,\lambda_3,\lambda_4$ are weighting factors. 

\subsection{Local Sketch Synthesis}
\label{ssec:local}

By carefully examining the globally synthesised portraits, we find eyebrows are not well generated occasionally. For light-colored eyebrows, the algorithm might produce no elements in this region. For eyebrows with non-uniform colors, the algorithm may produce a line with bifurcations. 

To address this problem, we additionally use a GAN to specifically synthesis eyebrows. Here we reproduce the local GAN in APDrawingGAN \cite{APDrawingGAN}. The local sketch synthesis network follows an U-Net architecture \cite{Pix2Pix} and trained on a small set of eyebrow photo-sketch pairs \cite{APDrawingGAN}. 

\subsection{Post-processing}
\label{ssec:postproc}
Finally, we implement a number of post-processing techniques to deal with the background and details. 

\textbf{Image Binarization.} The output of both the global and local synthesis algorithms are gray-scale sketches. We convert them to binary images based on a threshold of $0.5$. 

\textbf{Fusing Global and Local Sketches.}
Before fusion, we thin the locally synthesised eyebrows and remove delicate eyelashes \cite{APDrawingGAN}, by using image dilation.  
Afterwards, we replace the eyebrows in the globally synthesised portrait with these dilated eyebrows. 
We here use the face parsing mask $M$ to determine areas of eyebrows. 



\textbf{Eyeball Renewal.} 
Occasionally, our algorithm only produce a circle to represent an eyeball. Although this hardly harm the final sketch drawn by the robot, we propose a method to renewal an eyeball. Specially, we add a black spot to an eye center if it is blank. We determine the position of eye center based on facial key-points detection. 

\textbf{Style Fusion.} 
Given a facial picture, the global synthesis algorithm produces diverse sketches while using different style images. Such sketches might diverse in the number, width, and continuity of brush-strokes. It is therefore provide a chance to fuse these sketches based on the facial parsing mask. In our implementation, we use a style image to synthesis the primal sketch, and use another style image to synthesis the hair. Afterwards, we replace the hair region in the former sketch by those in the latter one. 


\section{The Robotic System}
\label{sec:robotic}
In this section, we introduce the hardware components and the path planning module, which implement the above synthesised portrait into a real sketch. 

\subsection{Hardware}
\label{ssec:hardware}

We use a commercial robotic arm (3 DOFs), i.e. uArm \cite{UArm}, to perform portrait drawing. 
We choose this robotic arm based on a trade-off between the performance and the cost. 
Using better robots might boost the performance, but will dramatically increase the cost for developing a portrait drawing robot. 
The workspace size on paper is about $160~\mathrm{mm} \times 160~\mathrm{mm}$. 
In this work, we have adopted writing brushes for portraits drawing. We integrate the brush with the robotic arm through plastic supports, which are manufactured by means of 3D printing.

To improve the mobility of the robot, we have implemented all the portrait synthesis and path planning algorithms on a remote server. 
In this way, the terminal of our drawing robot is composed of a robotic arm, a NVIDIA Jetson Nano board, and a mobile device.
The light weight and small volume of our robotic terminal dramatically ease its applications.
Besides, by using a NVIDIA GeForce GTX 1060 GPU on the remote server, we can transfer an input picture to a sketch in one second averagely. Such a high-speed synthesis meets the requirements of real-time applications.

\subsection{Path Planning}
\label{ssec:path}

Artists typically draw graphical lines along geometrical boundaries, and render black marks for eyebrows and eyeballs. To imitate this behaviour, we use the following methods to convert a synthesised portrait to a trajectory. 

We first extract the skeletons of a portrait image. The portrait typically contain brush-strokes with non-uniform width. It is time-consuming for a robot to elaborately rendering such details.
By extracting skeletons of brush-strokes, we reduce the elements for a robot to reproduce. We implement this by algorithms included in \texttt{OpenCV}. 
Afterwards, we search a sequence of points on skeletons, along the gradient orientations estimated by a Canny edge detector. Here breadth-first search is used. 
Finally, to fill the black marks representing eyeballs and eyebrows, we draw the closed loop of edge from outside to inside iteratively. 


\section{Experiments}
\label{sec:exp}
In this section, we first introduce implementation details, and then present a number of experiments to verify the performance of our portrait drawing robot.

\subsection{Implementation Details}
\label{ssec:implement}

\textbf{Datasets.} 
First, we need a large set of content images and a number of style images, to train the global sketch synthesis algorithm. To this end, we randomly select 1,000 low-resolution facial pictures from the CelebA dataset \cite{CelebA} and 1,000 high-resolution facial pictures from the CelebA-HQ dataset \cite{MaskGAN}, as the set of content images. All these pictures are resized into $512 \times 512$ pixels. We also download 20 real sketches of arbitrary content from the Web, as style images. We randomly choose 95\% of facial pictures for training, and use the rest for validation.  

Second, to train our local sketch synthesis algorithm, we extract 280 eyebrow photo-sketch pairs from the dataset released by \cite{APDrawingGAN}. Each sample is of $60 \times 60$ pixels. We randomly choose 210 samples for training, and the rest 70 samples for validation. 

Finally, we download about 200 images, including celebrity faces and universal images of arbitrary content, from the Web for testing. These facial pictures present faces in-the-wild, which are diverse in poses, expressions, etc. We apply our AiSketcher to these images so as to evaluate its performance in practical applications. The corresponding results have been released at: https://ricelll.github.io/AiSketcher/.

\textbf{Training.} 
In the training stage, we use the Adam optimizer with a learning rate of $1e-4$ and a batch size of $8$, for both the global and local sketch synthesis algorithms. 
Besides, to optimize the global algorithm, we have $\lambda_1 = \lambda_2 = \lambda_3 = 1$, and $\lambda_4 = 10$ in Eq.7, and run for 160,000 iterations.
To optimize the local algorithm, we alternate between one gradient descent step on $D$, then one step on $G$. We use the Adam optimizer with $\beta = 0.5$ and run for different $600$ epochs. 
We train our algorithms on a single NVIDIA GeForce GTX 1080Ti GPU. It takes about 22 hours and 1 hour to train the global and the local algorithms, respectively.

%


\subsection{Comparison with State-of-the-art}
\label{ssec:compare}

In this part, we compare our portrait synthesis algorithm with existing works about portrait drawing robots, i.e. \cite{Lin2009}, \cite{Xue2017}, and \cite{Gao2019}, and a state-of-the-art (SOTA) portrait synthesis method, i.e. APDrawingGAN \cite{APDrawingGAN}. For this purpose, we collect the pictures and synthesised portraits presented in \cite{Lin2009}, \cite{Xue2017}, and \cite{Gao2019}. Afterwards, we apply the officially released APDrawingGAN and our learned sketch synthesis algorithm to these facial pictures, respectively.
   
   	\begin{figure}[thpb]
      \centering
      \includegraphics[width=1\linewidth]{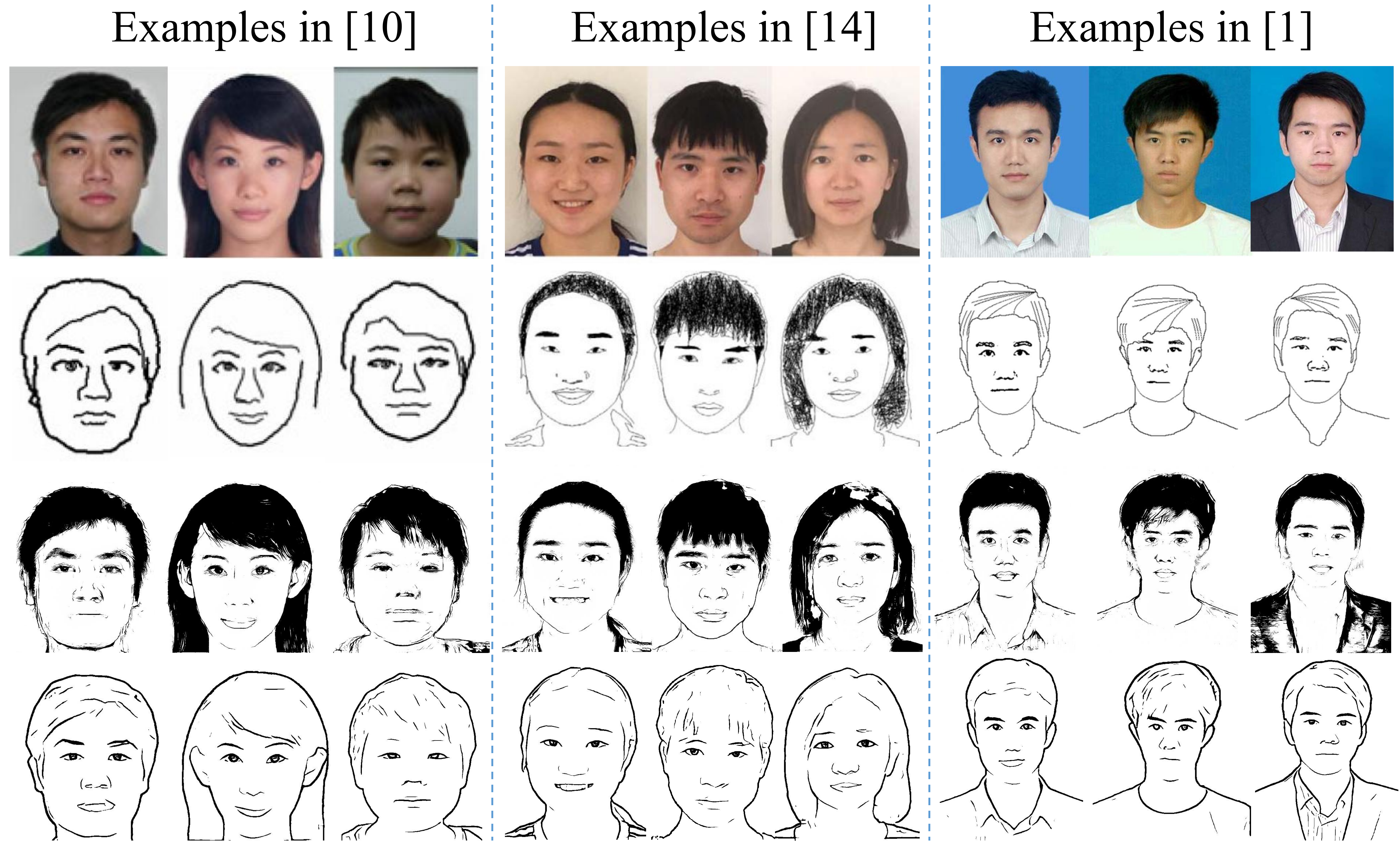}
      \caption{Comparison with existing portrait drawing methods. Input images are colected from \cite{Lin2009}, \cite{Xue2017}, and \cite{Gao2019}. The second row shows the corresponding results in the original papers. The third and bottom rows show sketches synthesised by APDrawingGAN \cite{APDrawingGAN} and our method, respectively.}
      \label{fig:compare}
   \end{figure}

As shown in Fig. \ref{fig:compare}, the portraits produced by \cite{Lin2009} are over simple and present sinuate lines. Portraits produced by both \cite{Xue2017} and \cite{Gao2019} represent facial structures by stiff and unrealistic lines. Besides, both of them cannot properly synthesis the hair. 
The portraits generated by APDrawingGAN capture the distinctive appearance of a person, but present low-quality elements. Besides, APDrawingGAN cannot produce portraits with a sparse set of continuous lines and marks. 

In contrast, the portraits synthesised by our algorithm successfully capture individual characteristics and feelings, by using a sparse set of realistic brush-strokes. Moreover, our algorithm produces a primal sketch for the area of cloths, which further improve the vividness of facial portraits. Note that these facial pictures are low-resolution and blurring. Results shown in Fig. \ref{fig:compare} demonstrate the remarkable capacity of our algorithm in generating high-quality portraits. 


\subsection{Qualitative Evaluation}
\label{ssec:exp_perform} 

\begin{figure}[thpb]
      \centering
      \includegraphics[width=1\linewidth]{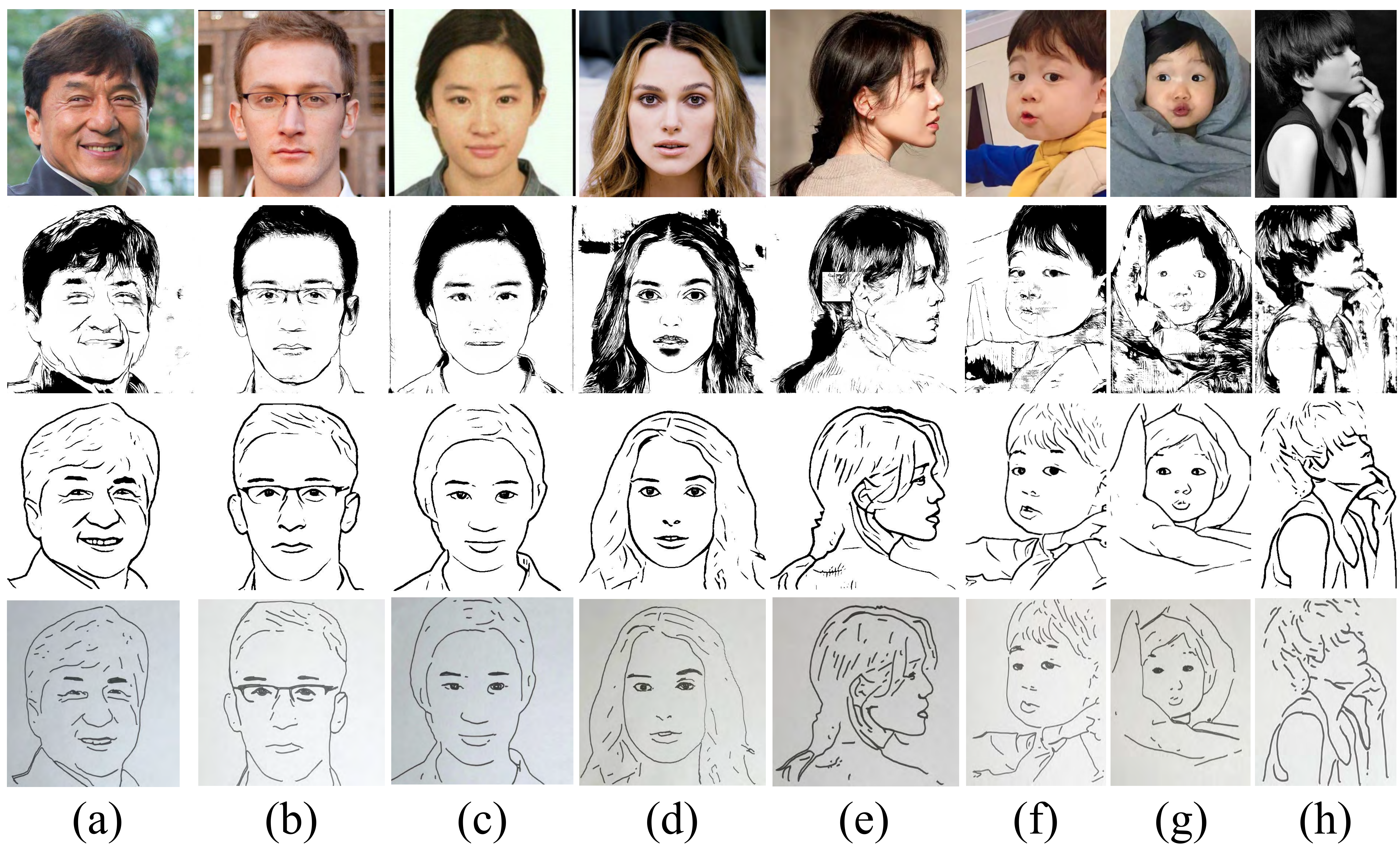}
      \caption{Sketchy portraits produced by our AiSketcher for national celebrities. From top to bottom are: input images, sketches synthesised by APDrawingGAN \cite{APDrawingGAN}, sketches synthesised and drawn by AiSketcher. All the input images shown here are downloaded from the Web. }
      \label{fig:example}
   \end{figure}

We further apply our robotic system to a wide range of faces in-the-wild. Such faces are diverse in illumination conditions, poses, expressions, and may contain occlusions. Fig. \ref{fig:example} show some examples. Here, we pre-process photos in Fig. \ref{fig:example}a-\ref{fig:example}d before feeding them into APDrawingGAN or our AiSketcher. While the photos in Fig. \ref{fig:example}e-\ref{fig:example}h are directly input into APDrawingGAN or our global synthesis algorithm for producing portraits. 

APDrawingGAN doesn't work properly for these photos. 
In contrast, our AiSketcher considerably and consistently produce high-quality sketches for all these photos. Our algorithm successfully deal with complex backgrounds (Fig. \ref{fig:example}a and 5b), glasses (Fig. \ref{fig:example}b), extreme poses (Fig. \ref{fig:example}e, 5f, and 5h), and occlusions (Fig. \ref{fig:example}f). Notably, given a half-length photo (Fig. \ref{fig:example}h), our algorithm is capable of producing a high-quality and abstract portrait. 

By comparing our synthesised sketches with the final versions drawn by AiSketcher, we conclude that the robotic arm precisely reproduces the synthesised sketches. Although some details are slightly missing, due to the limited capability of our 3-DOF robotic arm, the final sketches successfully present the individual appearances. 



\subsection{Generalization Ability}
\label{ssec:general} 
\begin{figure}[thpb]
      \centering
      \includegraphics[width=1\linewidth]{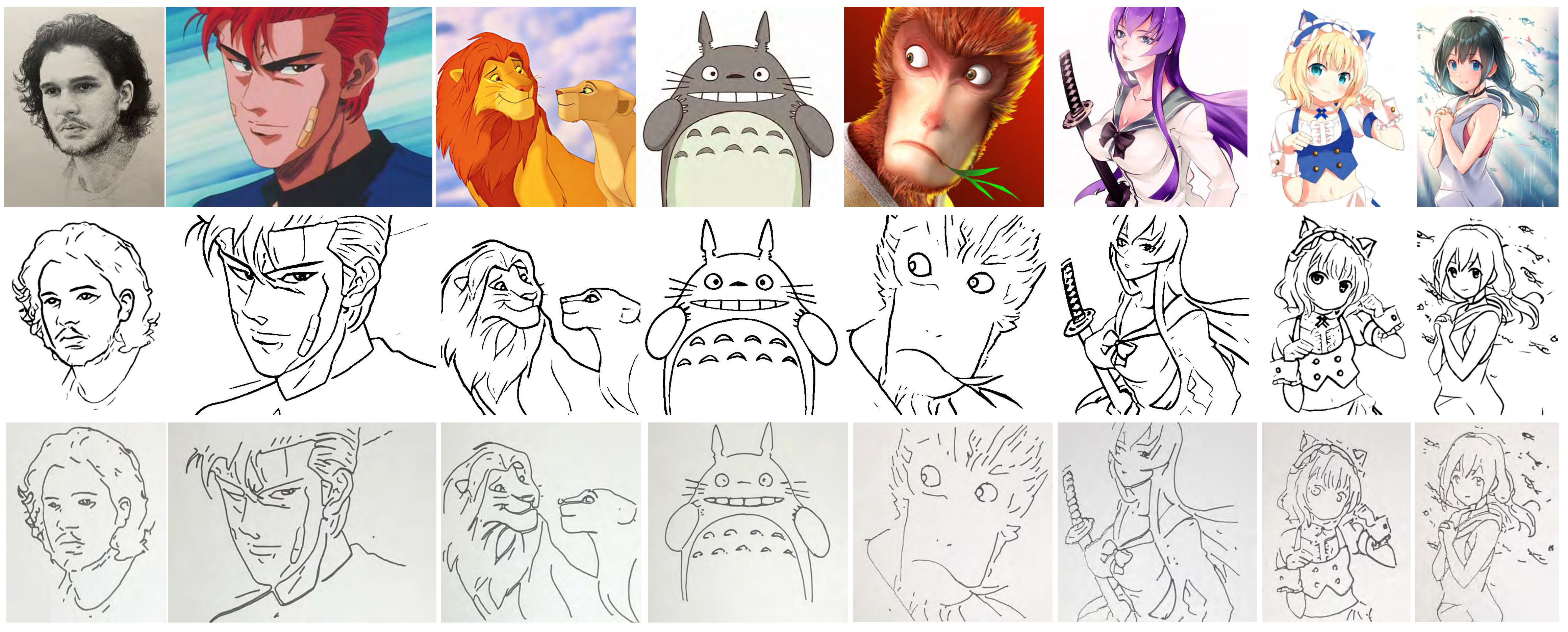}
      \caption{Synthesised sketches for universal images. From top to bottom are: input images, sketches synthesised and drawn by AiSketcher. All the input images shown here are downloaded from the Web. }
      \label{fig:universal}
   \end{figure}

We further apply AiSketcher to universal images, such as pencil-drawing images, cartoon images, etc. In this case, we DON'T use pre-processing, the local synthesis algorithm, or post-processing techniques (except image binarization), because there might be no human face in a given image. As illustrated in Fig. \ref{fig:universal}, AiSketcher still produce considerably high-quality sketches. A wide range of test shows that our AiSketcher tends to produce a high-quality sketch for an universal image, unless the image contains a large area of delicate textures or is low-quality (e.g. low-resolution, blurred, or noisy, etc.). 

\subsection{Time Consumption}
\label{ssec:timec}

In the testing stage, given an input picture, it takes about $1\mathrm{s}$ for AiSketcher to synthesis a sketch, by using a server with a NVIDIA GeForce GTX 1060 GPU. Afterwards, it costs about 1-3 minutes for the robotic arm to draw it. The total time-consuming is about 2 minutes in average. Thus AiSketcher fairly satisfy the time-budget in practical scenarios. Recall the considerable performance of AiSketcher in generating high-quality portraits, we conclude that AiSketcher successfully balance vividness and time-consumption.

\subsection{Ablation Study}
\label{ssec:ablation}

Finally, we present a series of ablation studies to verify the effects of our proposed techniques. For this purpose, we test several model variants derived from  our portrait synthesis algorithm, including: 
(i) the based model, AdaIN \cite{AdaIN}; (ii) AdaIN with the self-consistency loss, AdaIN w/ $L_{consist}$; (ii) AdaIN with both $L_{consist}$ and global sparsity, AdaIN w/ $L_{sparse}^{global}$; (iv) AdaIN with both $L_{consist}$ and compositional sparsity loss, AdaIN w/ $L_{sparse}$; and (v) our full model, including pre-processing, local synthesis, and post-processing. 

%

   \begin{figure}[thpb]
      \centering
      \includegraphics[width=1\linewidth]{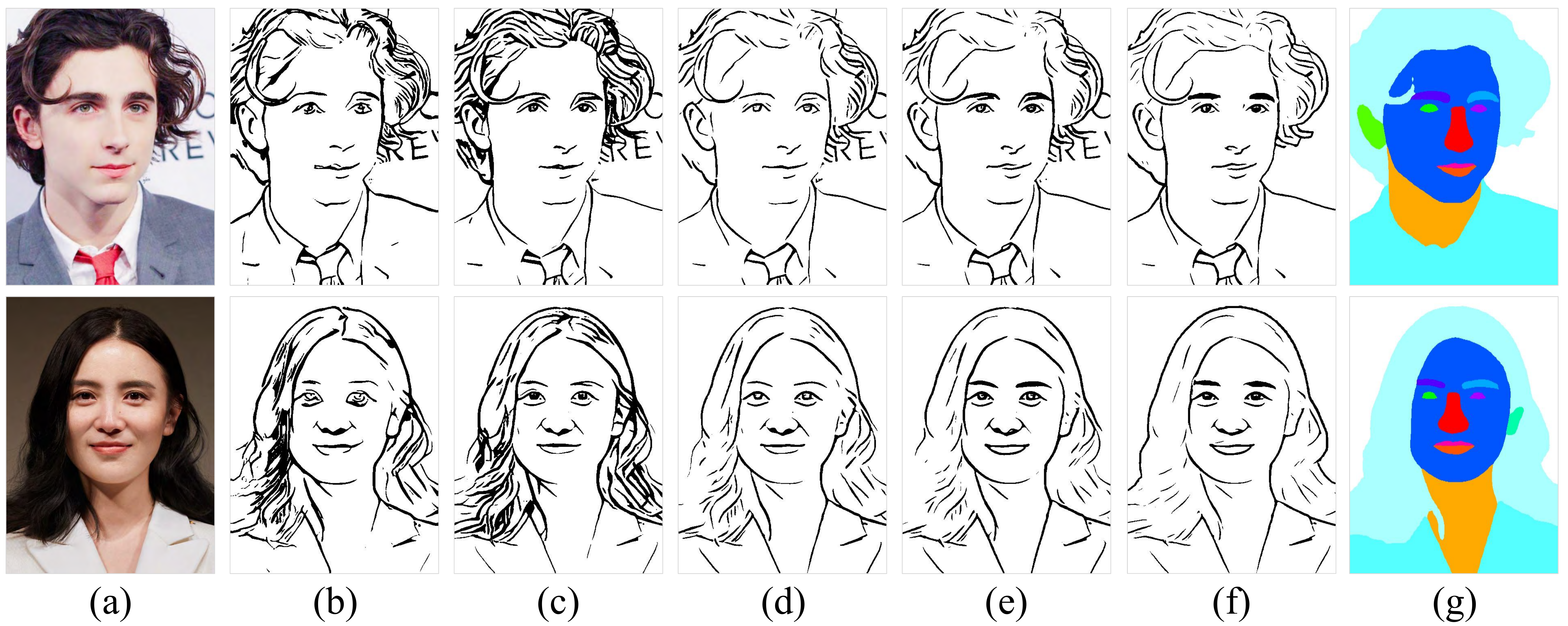}
      \caption{Ablation study on our novel portrait synthesis algorithm. From left to right are: (a) Input photo, and sketches synthesised by (b) AdaIN \cite{AdaIN}, (c) AdaIN w/ $L_{consist}$, (d) AdaIN w/ $L_{sparse}^{global}$, (e) AdaIN w/ $L_{sparse}$, and (f) our full model, (g) parsing mask predicted by using BiSeNet \cite{yu2018bisenet}.}
      \label{fig:ablation}
   \end{figure}
   
\textbf{Effects of Self-consistency Loss.} 
Fig. \ref{fig:ablation}a shows that AdaIN typically produces black marks over the hair region, and snatchy brush-strokes. In contrast, AdaIN w/ $L_{consist}$ produces smooth and continuous brush-strokes with few black marks. The apparent improvement demonstrates our motivation of using the self-consistency loss. 

\textbf{Effects of Compositional Sparsity.}
Sketches in Fig. \ref{fig:ablation}b present too many brushes, which requires a long drawing time. 
Both Fig. \ref{fig:ablation}c and \ref{fig:ablation}d show that using a sparsity constraint dramatically reduce the number of brush-strokes. 
However, using a global sparsity reduces the elements representing individual characteristics (e.g. the eyes and eyebrows in Fig.\ref{fig:ablation}c). In contrast, our compositional sparsity successfully avoid this problem. In other words, the compositional sparsity dramatically decrease the drawing time without apparently decreasing the quality of synthesised portraits. 

\textbf{Effects of Local Synthesis.} 
Fig.\ref{fig:ablation}d shows that the globally synthesised eyebrows may present lines with bifurcations. While replacing the eyebrows by the locally synthesised ones results in a better portrait, as shown in Fig.\ref{fig:ablation}e. 

\textbf{Full Model.} 
Fig.\ref{fig:ablation}f shows that the results produced by our full model, where local synthesis as well as pre- and post-processing techniques are applied. The resulting portraits present cleaner background and better eyebrows/eyeballs, compared to Fig.\ref{fig:ablation}e. 

Based on the above observations, we can safely draw the conclusion that our proposed techniques significantly improve the quality of synthesised portraits.


\section{CONCLUSIONS}
\label{sec:conclusion}

In this work, we present a portrait drawing robot, named AiSketcher, which shows fantastic performance over a wide range of images. Besides, experimental results demonstrate that AiSketcher achieves a balance between the quality and the drawing time of sketch portraits. 
By carefully examining the produced sketches by AiSketcher, we find there is still substantial room for improvement. First, our sketch synthesis algorithm typically produces undesired brush-strokes along boundaries of a shadow. Illumination normalization is a promising approach for solving this problem. 
Second, the robotic arm cannot precisely reproduce delicate elements in a synthesised sketch. 
It is interesting to optimize both the path planning and the portrait synthesis algorithms, by taking into account the limitations of a robot arm. Besides, visual feedback \cite{IROS2016Luo,IROS2016Berio} is another promising solution. Finally, we hope to further improve the drawing speed and boost the energy efficiency by developing robot-specific processors. Such processors are like Dadu \cite{Dadu} and Dadu-P \cite{Dadu-P}, where parallel architectures are designed for the robot's motion control and path planning, resulting in the first complete framework for robot-specific processing. We will explore these issues in the near future. 

\section*{Acknowledgements}
\label{sec:ack}
We acknowledge Prof. Yinhe Han for his contribution on the initial version of the drawing robot where AiSketcher originated.

\end{document}